\documentclass[10pt, a4paper]{article}
\usepackage{lrec-coling2024} 

\title{PAQA: Toward ProActive Open-Retrieval Question Answering}
\name{Pierre Erbacher$^{1}$, Jian-Yun Nie$^{2}$, Philippe Preux$^{3}$, Laure Soulier$^{1}$} 
\address{Sorbonne Université$^{1}$, University of Montréal$^{2}$, Université de Lille$^{3}$\\
         pierre.erbacher@isir.upmc.fr\\}

\abstract{
Conversational systems have made significant progress in generating natural language responses. However, their potential as conversational search systems is currently limited due to their passive role in the information-seeking process. One major limitation is the scarcity of datasets that provide labeled ambiguous questions along with a supporting corpus of documents and relevant clarifying questions. This work aims to tackle the challenge of generating relevant clarifying questions by taking into account the inherent ambiguities present in both user queries and documents. To achieve this, we propose PAQA, an extension to the existing AmbiNQ dataset, incorporating clarifying questions. We then evaluate various models and assess how passage retrieval impacts ambiguity detection and the generation of clarifying questions.
By addressing this gap in conversational search systems, we aim to provide additional supervision to enhance their active participation in the information-seeking process and provide users with more accurate results.
 \\ \newline \Keywords{clarifying question, ambiguity, conversational search} }

\begin{document}

\maketitleabstract

\section{Introduction}

Conversational systems like Bard or ChatGPT \citep{https://doi.org/10.48550/arxiv.2201.08239,https://doi.org/10.48550/arxiv.2208.03188,https://doi.org/10.48550/arxiv.2203.02155} excel at question answering, by generating fluent, long, comprehensive and factual responses thanks to human feedback and retrieval abilities. However, they fall short in conversational search tasks, lacking proactive participation in information seeking \citep{10.1145/3498366.3505816, DBLP:journals/corr/abs-2201-08808}. Indeed, these systems are mainly reactive and trained to passively respond to user queries without actively participating in the conversation or understanding user needs. Due to the inherent ambiguity in natural language, these reactive systems might fail to capture the intent behind queries, especially when multiple interpretations exist in retrieved documents \citep{10.1145/3020165.3020183,DBLP:journals/corr/abs-2201-08808}. A key to successful search sessions is to proactively engage with users to clarify the queries. Current strategies involve asking clarifying questions from the system side, a critical aspect of query disambiguation \citep{Aliannejadi19-askingclarifying, Zamani-Clarification}.
Initial methods for selecting clarifying questions, such as those proposed by \citet{Aliannejadi19-askingclarifying}, utilize iterative ranking from a fixed pool, a tactic expanded upon by \citet{10.1145/3471158.3472232} utilizing negative feedback to better rank questions. However, these fixed-question based systems face topic coverage constraints. Alternative generative methods \cite{10.1007/978-3-030-72113-8_39}  use templates and topics collected from the Autosuggest Bing and later refined for fluency using a Language Model (LM) model by \cite{10.1145/3471158.3472257}. However, they still generate straightforward and single-topic clarifying questions of the type 'Are you interested in \textit{[topic]}?". \citet{xu-etal-2019-asking} utilize a knowledge base (KB) to generate questions, targeting only entity-based ambiguities with common names. \citet{Zamani-Clarification}'s strategy involves multi-faceted questions using query reformulation data, but the absence of actual documents in the MIMICS datasets limits their scope. These generative models depend heavily on extracting keywords from retrieved documents, underscoring the challenge posed by ambiguities that lie in both the questions and documents.
\begin{figure}[t]
    \centering
    \includegraphics[width=0.50\textwidth]{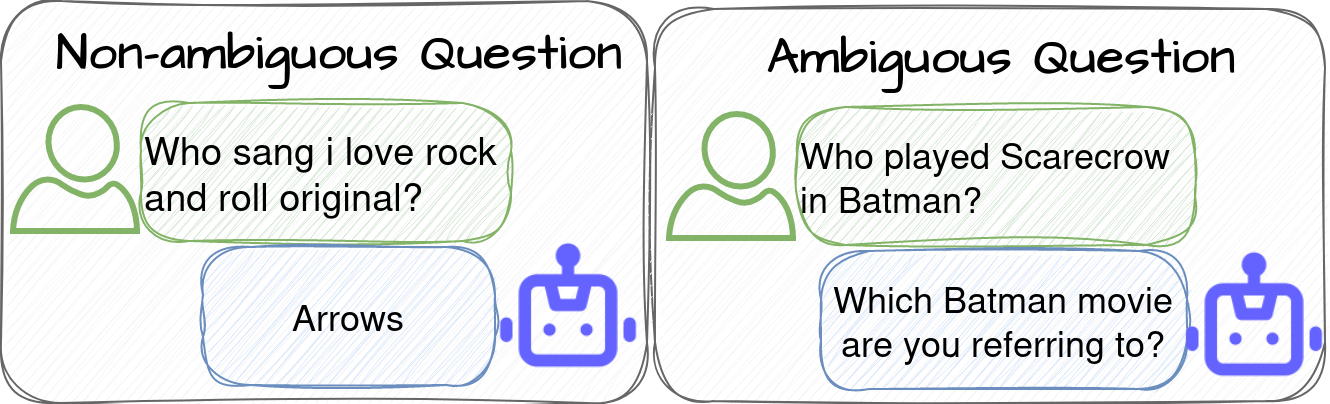}
    \caption{The figure depicts the proactive question-answering task. The system answers directly for non-ambiguous questions (left) and ask clarifying questions for ambiguous questions (right)}
    \label{fig:task}
\vspace{-0.25cm}
\end{figure}
Identifying possible underlying ambiguities requires models to discern probable answers across multiple documents.
This paper addresses these gaps by proposing the PAQA dataset containing annotated ambiguous questions with clarifying questions resolving underlying ambiguities and annotated documents, aiming for improved training and evaluation of systems in recognizing and resolving ambiguities. The dataset is built upon the AmbiNQ \cite{min-etal-2020-ambigqa} dataset by providing clarifying questions generated with GPT3. The dataset is supplied with reference models aiming at jointly detecting ambiguity and generating clarifying questions when necessary. We carried out human and automatic experiments to assess the relevance of the built dataset. 
The dataset is accessible at the following repository\footnote{\label{note1}\url{https://anonymous.4open.science/r/PAQA-dataset-7932/README.md}}.

\vspace{-0.2cm}
\section{Task Definition}
\label{task}
\vspace{-0.2cm}
A typical open-retrieval question-answering task involves extracting information from large collections of documents to answer accurately to questions.
The proactive question-answering task, as illustrated in figure \ref{fig:task}, additionally focuses on detecting and resolving ambiguities in questions posed to a system. The system is faced with two types of questions: non-ambiguous ones for which the system can directly answer and ambiguous ones for which the system must recognize that there are multiple possible interpretations. 
In this last case, instead of providing a direct answer that may not be relevant, the system aims to clarify the user's need by posing a clarifying question. These interactions assist users in specifying their needs, allowing systems to deliver more precise answers in subsequent conversation turns.
Formally, questions $q$ are associated with a set of equally plausible answers $a_1, ..., a_n$ grounded to a collection of passages $p\in P$, and considered ambiguous if $n > 1$. The system is expected to output the answer directly when the question is not ambiguous $(n=1)$ and to ask a clarifying question $cq$ that resolves the underlying ambiguity otherwise. Systems should be evaluated on their ability to detect ambiguities and to generate relevant clarifying questions.

 

\vspace{-0.1cm}
\section{PAQA dataset}
\label{paqa}
\vspace{-0.1cm}
Training and evaluating models to perform the task described in section \ref{task} requires having a large dataset with labeled ambiguous questions, annotated documents within a collection, clarifying questions and answers. Because there are currently no known datasets providing this supervision, we propose the PAQA dataset, an extension of the existing AmbigNQ datasets  that additionally contains clarifying questions (section \ref{ambignq}). We also provide reference models (Section 3.2) and evaluation metrics (Section 3.3). 
\subsection{Extending AmbigNQ with clarifying questions}
\begin{figure}[t]
    \centering
    \includegraphics[width=0.55\textwidth]{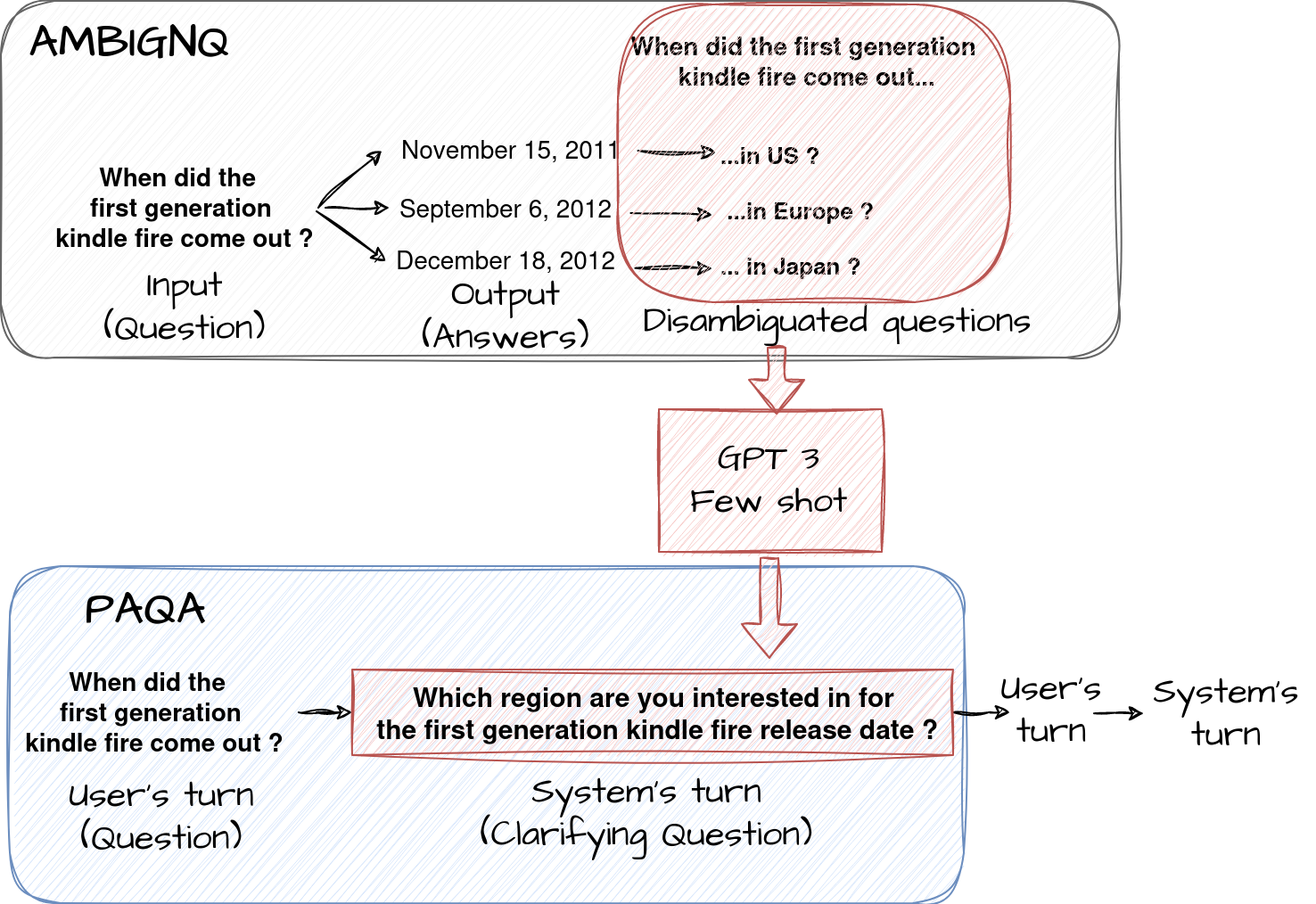}
    \caption{PAQA dataset: enriching the AmbigNQ dataset with clarifying questions.}
    \label{fig:methodology}
\end{figure}

\paragraph{Source Collection}
\label{ambignq}
The AmbigNQ \citep{min-etal-2020-ambigqa} dataset is a question-answering dataset specifically designed to tackle ambiguous questions in an open-domain setting. It was introduced with the AmbigQA task which consists in predicting a complete set of plausible answers given ambiguous questions. This dataset is constructed on NQ-open \citep{kwiatkowski2019natural} and contains about 14000 manually annotated examples across various topics.  Questions $q$ are associated with multiple pairs of plausible query interpretations and corresponding answers $(q_1,a_1),..., (q_n,a_n)$ with $n \ge 1 $ found on Wikipedia. Questions are considered non-ambiguous if $n=1$. One example is provided in the top of Figure \ref{fig:methodology} (the AmbigNQ grey box). The dataset is balanced between ambiguous and non-ambiguous questions. Additionally, to the Wikipedia dump containing (21 millions passages), AmbigNQ also provides semi-oracle (evidence) articles, which are 3 annotated Wikipedia pages either containing answers or not (see \citep{min-etal-2020-ambigqa}). AmbigQA is associated with 2 tasks: 1) multiple answers prediction: Given $q$, output the set of plausible answers $(a_1,..a_n)$ with $n$ being unknown. And 2) Question Disambiguation that given $q$ and the set of answers $(a1, ...,an)$ output the $(q1, ...,qn)$ with minimal edit. In the context of conversational search between a system and a user, both AmbigQA's tasks are not suitable, as the goal is to cover all answers without considering the user's information needs. Thus, motivation to augment the dataset with clarifying questions to allow proactive clarification.

\paragraph{Methodology}
As illustrated in Figure \ref{fig:methodology}, given the set of disambiguated questions $(q_1, ... q_n)$, we generate a clarifying question $cq$ asking how the question $q$ should be interpreted.  We used openai's GPT3-davinci \footnote{https://openai.com/} with few-shot examples. The prompt is structured as follows: 
\begingroup
\scriptsize  
\begin{verbatim}
"Generate a clarifying question given 
multiple queries \n\n"
    "Examples:\n"
    "----------\n"
    "Query1:Who is the 4th chairman of 
    african union commission?\n"
    "Query2:Who is the 3rd chairman of 
    african union commission?\n"
    "Query3:Who is the 2nd chairman of 
    african union commission?\n"
    "Question:Are you interested in 2nd,
    the 3rd or the 4th chairman of
    african union commission \n"
    "----------\n"
    ...[other examples]
    "----------\n"
    "Query1:How many teams are there in
    the afl in 1960-1965?\n"  
    "Query2:How many teams are there in
    the afl in 1966-1967?\n"
    "Query3:How many teams are there in
    the afl in 1968–1970?\n"
    "Question:" 
\end{verbatim}
\endgroup

The resulting  PAQA\footref{note1} dataset contains questions, associated sets of possible answers, annotated documents containing answers (evidence), clarifying questions, and the Wikipedia collection. Train/Dev/test sets contain, respectively, 10000/1000/1000 data points balanced with   $47.1\%$/$54.2\%$/$52.5\%$ of ambiguous questions, with a mean of 2.9/3.3/3.2 answers for ambiguous questions.

\subsection{Evaluating generated questions}
We evaluate the quality of the clarifying questions in PAQA using human evaluators. 100 clarifying questions and their associated set of disambiguated questions are sampled from the dataset, and 6 participants are asked to assess both Naturalness and Relevance. Naturalness is defined as being written in a fluid manner, in everyday language, coherent, and could have been generated by a human. Annotators can choose between Unnatural, Fair, Good, and Natural. Relevance measures how the clarifying questions cover/represent the set of provided disambiguated questions. Annotators can choose between irrelevant, partially relevant, and fully relevant. Each question is annotated by 2 humans. It is important to note that our annotators, despite not being specialists in the field or native speakers of English, are well-equipped for their task. The assignment is straightforward and evaluators are guided by comprehensive guidelines and illustrative examples, that ensures clarity. Evaluators are presented with the ambiguous queries, corresponding answers, disambiguated questions from the AmbigNQ dataset, and the automatically generated clarifying questions. They assess the relevance by comparing the scope of answers covered in the clarifying questions compared with the 'disambiguated queries' annotated in the AmbigNQ dataset. 

\begin{table*}[t]
\resizebox{\linewidth}{!}{
\begin{tabular}{lp{14cm}}
\hline
Question                        & When did england last make the quarter final of the world cup?                                                                                   \\
Label                         & Are you looking for the men's or women's FIFA World Cup?                                                                                         \\ \hline
$Q$                & Are you looking for the last time England made the quarter final of the world cup or the last time they made the quarter final of the world cup? \\ \hline
$Q$ + $P_{evidence}$            & Are you looking for the men's or women's world cup quarter finals?                                                                               \\
$Q$ + $P_{evidence}$ + $A_{gold}$ & Are you looking for the men's or women's world cup quarter finals?                                                                               \\ \hline
$Q$ + $P_{dpr}$             & Are you looking for the quarter final of the FIFA World Cup or the Rugby World Cup?                                                              \\
$Q$ + $P_{dpr}$  + $A_{extracted}$     & Do you want to know the year or the date of England's quarter final appearance in the FIFA World Cup?                                            \\ \hline

\end{tabular}
}
\caption{Table showing qualitative examples of generated cq}
\label{table:qualitative}
\vspace{-0.25cm}

\end{table*}

\vspace{-0.25cm}
\subsection{Proactive question answering with generative model}

\paragraph{Model architecture}
While the task can be decomposed into a classification task and a generation task, our approach involves using a single generative model to tackle the whole task by generating either the answer or a clarifying question when needed. We consider a sequence-to-sequence architecture trained to generate a clarifying question $cq$ for input question $q$ associated with multiple plausible answers and directly provide answers if the question $q$ is non-ambiguous. We explore different versions of the model: \\
\textbf{Question-only $Q$}:
The model is only conditioned on the input by the question $q$  during training and inference. This means that the LM must rely entirely on parametric knowledge for this task.\\
\textbf{Retrieval Augmented $Q+P_{evidence}$ or $Q+P_{dpr}$}: The clarifying model takes as input a set of $n$ passages $p_1...p_n$ in addition to the query $q$.  Passages noted $P_{dpr}$ are retrieved by a dense passage retriever (dpr) model \cite{karpukhin-etal-2020-dense} and re-ranked with a cross-encoder. Passages noted $P_{evidence}$ are evidence passages included in the AmbigNQ dataset (section \ref{ambignq}). \\   
\textbf{Retrieval Augmented with Pre-extracted Answers}:
 This setting refers to the retrieval augmented model by integrating answers extracted from the retrieved passages. To do so, given top-n passages $p_1...p_n$, we use a QA model to predict answers $a_1...a_n$ associated with scores $s_1...s_n$. Answers are filtered to maintain a set of unique answers, and we consider a threshold to filter predicted answers based on their scores. Answers are denoted $A$ in the results. Answers noted $A_{gold}$ and $A_{extracted}$ are gold and automatically extracted answers, respectively. \\ 
 For all model variants, inputs (if considered) are concatenated as follows: $Question: q Docs: p1..p_n Answers: a1,..,a_n$.

\paragraph{Implementation details}
We implemented our code base using the Transformers library \citep{wolf-etal-2020-transformers}. For the generative model, we used the Flan-Large \citep{chung2022scaling} available on the Huggingface hub\footnote{https://huggingface.co/google/flan-t5-large}. This model has a context size of 512 tokens and can take 4 passages as input. We used the pre-trained dense retrieval model DPR \citep{karpukhin-etal-2020-dense} to encode questions and retrieve Wikipedia passages. To train the seq-to-seq model, we follow  \citet{min-etal-2020-ambigqa} and split evidence into 100-word passages, and re-rank them using a cross-encoder given the question q. For re-ranking passages, we rely on the pre-trained MonoT5\footnote{https://huggingface.co/castorini/monot5-base-msmarco-10k} \citep{Pradeep-monoduo} trained on MSMARCO \citep{NguyenRSGTMD16-MsMarco}. For Automatic answers extraction, we rely on the deberta-large model \citep{he2020deberta} trained on squad \citep{rajpurkar-etal-2016-squad}. We used a learning rate of $2e-5$ with a batch size of $64$ for $15$ epochs. 

\subsection{Automatic Evaluation Metrics}
Models evaluated on the PAQA task should be evaluated on their ability to detect underlying ambiguities and to generate relevant clarifying questions.  The F1 score, recall, and precision are known metrics to evaluate classification performances. The quality of generated clarifying questions can be evaluated using  ROUGE \citep{lin-2004-rouge}, METEOR \citep{banerjee-lavie-2005-meteor}, BLEU \citep{papineni-etal-2002-bleu} commonly used for generative tasks with available gold references.

\vspace{-0.2cm}
\section{Results}
\vspace{-0.2cm}
\begin{table}[t]
\resizebox{\linewidth}{!}{
\begin{tabular}{lccc}
\hline
\multicolumn{1}{c}{}                         & ROUGE-L         & BLEU-1           & METEOR   \\ \hline
Q                                     & 0.512          & 0.180          & 0.480           \\ \hline
Q + $P_{evidence}$                         & 0.554          & 0.213          & 0.518          \\
Q + $P_{evidence}$ + $A_{gold}$                  & \textbf{0.560} & \textbf{0.225} & \textbf{0.527}  \\ \hline
Q + $P_{dpr}$           & 0.546       & 0.204       & 0.509      \\
Q + $P_{dpr}$  + $A_{extracted}$ & 0.519          & 0.189          & 0.484        \\ \hline
\end{tabular}
}
\caption{Table showing performance for generating clarifying questions with various inputs. Metrics are computed only on generated cq.}
\label{table:generation}
\vspace{-0.3cm}
\end{table}

\begin{table}[t]
\resizebox{\linewidth}{!}{
\begin{tabular}{lccc}
\hline
\multicolumn{1}{c}{}                         & Accuracy          & Precision           & Recall   \\ \hline
Q                                     & 0.527          & 0.535          & 0.920           \\ \hline
Q + $P_{evidence}$                           & 0.644          & 0.699          & 0.569          \\
Q + $P_{evidence}$ + $A_{gold}$                   & \textbf{0.873} & \textbf{0.952} & \textbf{0.798}  \\ \hline
Q + $P_{dpr}$          & 0.572       & 0.619       & 0.487      \\
Q + $P_{dpr}$  + $A_{extracted}$  & 0.565       & 0.621   & 0.447        \\ \hline
\end{tabular}
}
\caption{Table showing performance for classifying ambiguous questions with various inputs. }
\label{table:classification}
\vspace{-0.5cm}
\end{table}

\subsection{Human evaluation results}


Our sample of evaluated questions is considered mostly natural with 0 Unnatural, 1 Fair, 5 Good, and 94 Natural. For relevance: 1 is unrelated, 18 are partially relevant meaning that they are not covering all the intents, and 81 are fully relevant covering all the intents. The consensus among annotators, as reflected by Cohen's kappa scores, stands at 0.68 for the naturalness and 0.79 for the relevance of the questions. This indicates a high level of agreement on both counts. Concerning the 18 Partially Relevant, we observed that 17 of them have 6 or more intents. When there are numerous intents, the language model tends to group or leave suggestions in order to generate more natural questions, which is more desirable than list-like generation, especially for natural conversational applications where multiple rounds of clarification are required.
The unrelated is indicating a likely false negative, given the intents: \textit{Japanese hotel run by the same family for 1300 years in (Komatsu?/Hayakawa?)} the generated query, \textit{Are you looking for a hotel in Komatsu or Hayakawa?}, seems unrelated. Its phrasing implies a search for booking instead of specifics about the centuries-old family-run Japanese hotel. Overall, the human evaluation shows that clarifying questions are very natural and relevant regarding annotated ambiguities.




\vspace{-0.2cm}
\subsection{Effectiveness of baselines}
\vspace{-0.2cm}
Tables \ref{table:generation} and \ref{table:classification} showcase generation and classification results across various configurations. Models using evidence ($P_{evidence}$) outperform retrieval pipelines in accuracy and generation with an accuracy of $0.64$ with evidence and $0.57$ with passaged retrieved from DPR. This suggests that the retrieval quality highly affects performances. The model relying on parametric knowledge ("Question only") scores lower in BLEU, ROUGE, and METEOR due to the lack of context and produces poor clarifying questions as seen in Table \ref{table:qualitative}.
Surprisingly, pre-extracting answers do not yield better results and marginally lower the generation and accuracy metrics, from a BLEU of $0.20$ for the DPR setting to $0.18$ with the model relying on DPR and answers. This is probably because the extractive QA predicts non-relevant answers with high probability, penalizing the entire pipeline.
An example is shown in table \ref{table:qualitative}, where both models with evidence manage to match the labeled CQ, the model with DPR only manages to generate another clarifying question, and the DPR + answer is influenced by retrieved answers with different date formats. The results show that PAQA enables models to generate accurate clarifying questions and detect ambiguities, however, the main bottleneck remains in the retrieval capabilities with a performance gap between models with evidence and retrieved passages. 

\vspace{-0.3cm}
\section{Conclusion}
\vspace{-0.2cm}
While conversational systems have made significant progress in generating natural language responses, 
they do not ask relevant clarifying questions when faced with documents supporting multiple plausible answers but rather generate a comprehensive answer. We propose a new dataset with aligned documents, questions, and clarifying questions to better train and evaluate such systems in their ability to detect ambiguity and ask clarifying questions.
To our knowledge, PAQA is the first dataset providing both questions, answers, supporting documents, and clarifying questions covering multiple types of ambiguity (entity references, event references, properties, time-dependent…) with enough examples for fine-tuning models. We provide various baselines with generative models. Experiments suggest that retrieval quality highly affects clarification quality and that pre-extracting answers do not lead to better ambiguity detection.  We did not investigate how scaling model parameters, nor how increasing the number of retrieved passages influences performances.
In a system, this may be preferable to have multiple rounds of partially-relevant but natural clarifying questions rather than a single clarifying question, increasing the cognitive load of users by listing all possible interpretations. Please note that we let for future work this perspective of building multiple rounds of clarifying questions as this requires simulating users' answers.


\section*{Ethics Statement}
In this paper, we do not use any sensitive, proprietary data, or personal data. Natural Question was collected using crowed sourced workers, documents used are from  Wikipedia. We also used Flan-T5, which is freely available. PAQA will be released to facilitate reproducibility.
However, we used few-shot on proprietary model to generate the supervision. This may include bias in the way clarifying questions are formulated. 
This research could be used to include at lower cost supervised examples of clarifying questions in conversational data, however, because this data is synthetic this may not be aligned with human preferences.

Furthermore, this raises ethical concerns about the carbon footprint associated with training and testing Large Language Models. This works aims at improving LLM abilities to engage and clarify user's need, potentially increasing the number of interactions for a single information need. This means increasing carbon emissions per search session.

\section{Data and Code statement}
The dataset is  available and can be accessed at anonymized repository \url{https://anonymous.4open.science/r/PAQA-dataset-7932/README.md} 
 

\appendix

\section{Appendix: PAQA description}
\begin{table}[h]
\begin{tabular}{lllll}
\hline
      & Size  & Ambig &  n & length cq \\
train & 10000 & 47.1\%    & 2.9            & 18.5       \\
dev   & 1001  & 54.2\%    & 3.3            & 18.6       \\
test  & 1000  & 52.5\%    & 3.2            & 18.5       \\ \hline
\end{tabular}
\caption{Table Describing the PAQA dataset. $n$ is the average number of answers for ambiguous questions. Length cq is the mean length of generated clarifying questions. Ambig the percentage of ambiguous question.  }
\label{tab:data_description}
\vspace{-0.5cm}
\end{table}

\begin{table*}[]
\begin{tabular}{l}
\hline
\begin{tabular}[c]{@{}l@{}}\textbf{Q}:When did the first generation kindle fire come out?\\
\textbf{Intents, Answers:}\\ 1. When did the first generation kindle fire come out in US?, A:November 15, 2011\\ 
2. When did the first generation kindle fire come out in Europe?, A:September 6, 2012\\ 
3. When did the first generation kindle fire come out in Japan?, A:December 18, 2012 \\
\textbf{CQ}:Which region are you interested in for the first generation kindle fire release date?\\ 
\end{tabular}                                                                                          \\ \hline
\begin{tabular}[c]{@{}l@{}}
\textbf{Q}:How many times has green bay beat the bears?\\ 

\textbf{Intents, Answers:}\\  
1. How many times has green bay beat the bears in the 2010s?, A: 17\\ 
2. How many times has green bay beat the bears in the 2000s?, A:12\\ 
3. How many times has green bay beat the bears in the 1990s?, A:13\\ 
4. How many times has green bay beat the bears all time?, A:99\\
5. How many times has green bay beat the bears consecutively?, A:10\end{tabular} \\ 
\textbf{CQ}:Do you want to know about the 2010s, 2000s, 1990s or all time, and are you interested in consecutive wins?\\ \hline
\end{tabular}
\caption{Table showing example from PAQA. Q is the original question, plausible intents and answers are labelled by annotators in the origial AmbigQA dataset. CQ are  clarifying questions generated using GPT3.}
\label{tab:ambigQA-cq}
\end{table*}
\section{Appendix: Human evaluation metrics}
\begin{table*}[h]
\begin{tabular}{l p{13cm}}
\hline
          & \multicolumn{1}{c}{Naturalness}                                                                                            \\ \hline
Unnatural & Sounds awkward or incorrect, either in terms of grammar, vocabulary, or syntax                                             \\
Fair      & Understandable, but still has some issues with flow or clarity                                                             \\
Good      & well-written, clear, and flows nicely, but may not necessarily sound completely natural or effortless                      \\
Natural   & well-written and sounds like something a native speaker would say, with appropriate vocabulary and syntax for the context. \\ \hline
\end{tabular}
\caption{Table describing the metrics description used to annotate naturalness of clarifying questions}
\label{table:naturalness}
\end{table*}
\label{sec:appendix}

\section{Appendix: Additional Examples}

\begin{table*}[h]
\begin{tabular}{ll}
\hline
          & \multicolumn{1}{c}{Relevance}                                                                                            \\ \hline
Irrelevant & Not related to the question or contains hallucinations                                             \\
Partially-Relevant      & Relevant but does not cover all the possible interpretations $Q1,...Qn$                                                             \\
Fully-Relevant      & Relevant and covers all annotated intents $Q1,...Qn$              \\
 \hline
\end{tabular}
\caption{Table describing the metrics description used to annotate relevance of clarifying questions }
\label{table:relevant}
\end{table*}


\begin{table*}[t]
\begin{tabular}{lp{14cm}}
\hline
Question                        &Who wrote the music for how to train your dragon?                                                                                   \\
Label  & Are you looking for the composer or the lyricist of the song "Sticks and Stones" for the film How to Train Your Dragon?                                                                                         \\ \hline
Q only                & Are you looking for the composer or the composers of the music for how to train your dragon? \\
Q+E            &       Are you looking for the composer of the music for How to Train Your Dragon, How to Train Your Dragon 2 or How to Train Your Dragon: The Hidden World?                                                                       \\
Q+E+A & Do you want to know about the music for the film or the song "Sticks \& Stones"?                                                                               \\
Q+P     & Do you want to know about the music for the film, the soundtrack album, or the song "How to Train Your Dragon?                                                              \\
Q+P+A     & Are you looking for the composer of the music for the film or the soundtrack album for How to Train Your Dragon?                                           \\ \hline  \hline

Question       &  Who wrote the books of luke and acts?                                                                                 \\
Label  &          Do you want to know about the traditional view or the critical views?                                                                               \\ \hline
Q only                & Are you looking for the author of Luke or Acts? \\
Q+E            &         Do you want to know who wrote the books of luke and acts in general or who specifically wrote the books of luke and acts?                                                               \\
Q+E+A &      Are you looking for the traditional or critical view of who wrote the books of luke and acts?                                                                          \\
Q+P     &   Do you want to know who is traditionally believed to have written the books of luke and acts or who modern scholars believe wrote the books of luke and acts?                                                       \\
Q+P+A     & Do you want to know who wrote Luke or Acts?
\\ \hline  \hline

Question       &  When did the movie inside out come out?
                                                                                 \\
Label  &    Are you looking for the release date of the 2015 movie inside out at Cannes, throughout the United States, at the El Capitan Theatre, in limited New York City theatres, or for a different year?                                                                             \\ \hline
Q only                & Are you looking for the release date of Inside Out in the United States or Canada? \\
Q+E            &    Do you want to know about the 1986, 2011 or 2015 movie Inside Out?                                                                   \\
Q+E+A &     Do you want to know about the 2015, 1986 or 1975 movie Inside Out?                                                                          \\
Q+P     &   Are you looking for the release date of the 2015 or the 1975 movie Inside Out?                                                     \\
Q+P+A     & Are you looking for the release date of the movie inside out in competition at the Heartland Film Festival, limited theatrical release, or on DVD and VOD?
\\ \hline  \hline

\end{tabular}
\caption{Table showing qualitative examples of generated cq}
\label{table:more_qualitative}
\end{table*}

\subsection{Extra space for ethical considerations and limitations}

\section{Bibliographical References}\label{sec:reference}

\bibliographystyle{lrec-coling2024-natbib}
\bibliography{custom}

\label{lr:ref}
\bibliographystylelanguageresource{lrec-coling2024-natbib}
\bibliographylanguageresource{languageresource}

\end{document}